%% file: main.tex
\documentclass{article}
\usepackage{iclr2027_conference,times}
\usepackage[T1]{fontenc}
\usepackage[utf8]{inputenc}
\usepackage{amsmath,amssymb,amsthm,graphicx,booktabs}
\usepackage{xspace}
\usepackage{xcolor}
\usepackage{colortbl}
\usepackage{makecell}
\usepackage{multirow}
\usepackage{pifont}
\usepackage{enumitem}
\usepackage{adjustbox}
\usepackage{xurl}
\usepackage[hidelinks]{hyperref}

\definecolor{shadecolor}{rgb}{0.92,0.92,0.92}

\usepackage{tcolorbox}
\tcbuselibrary{skins, breakable}

\newtcolorbox[auto counter]{findingbox}{
    breakable,
    enhanced,
    before skip=3pt,          
    after skip=3pt,
    top=2pt,                  
    bottom=2pt,
    left=3pt,                 
    right=3pt,
    pad at break=1pt,         
    bottomsep at break=0pt,   
    colback=gray!10,          
    colframe=black,           
    boxrule=1.1pt,            
    sharp corners,            
    fontupper=\itshape,       
    title={Finding \thetcbcounter.},
    colbacktitle=gray!10,
    coltitle=black,
    fonttitle=\bfseries\upshape,
    attach title to upper,
    after title={\hspace{0.5em}} 
}

\newcommand{\hi}[1]{\vspace{.25em} \noindent {\bf #1} }

\newcommand{\bfit}[1]{\textbf{\textit{#1}}}
\newcommand{\oursys}{\texttt{CRAFT}\xspace}
\newcommand{\ourbench}{\texttt{FormFillBench}\xspace}

\newcommand{\cmark}{\textcolor{green!60!black}{\ding{51}}}
\newcommand{\xmark}{\textcolor{red!75!black}{\ding{55}}}
\newcommand{\pmark}{\textcolor{orange!85!black}{$\blacktriangle$}}

\usepackage{bm}

\theoremstyle{definition}
\newtheorem{definition}{Definition}[section]
\newtheorem{example}{Example}

\title{CRAFT: An Agentic Spreadsheet Form Filling System with Template Awareness}
\author{
Leyao Gu$^{1}$ \quad
Yingjie Xiong$^{1}$ \quad
Zirui Tang$^{1}$ \quad
Jiangtao Zhou$^{1}$ \quad
Yeye He$^{2}$ \quad
Chunwei Liu$^{3}$ \\
\textbf{Xuanhe Zhou$^{1}$ \quad
Fan Wu$^{1}$} \\
$^{1}$Shanghai Jiao Tong University \qquad
$^{2}$Microsoft Research \qquad
$^{3}$Purdue University
}

\iclrfinalcopy

\begin{document}
\maketitle
\begin{abstract}
Spreadsheet form filling requires agents to consolidate external evidence, ground values to precise cells, and preserve irregular template structure. Errors in early edits can overwrite labels or misalign fields, undermining later decisions. We propose \oursys, a template-aware agent framework that connects reflective validation to constrained local repair. Instead of treating reflection as a free-form request to regenerate the workbook, \oursys grounds detected errors to spreadsheet regions, restores corrupted template state when necessary, and re-grounds plausible writable slots before subsequent edits. A Rectangle-Aware Slot Grounder (RASG) proposes writable cells, while label--slot hints and protected regions constrain subsequent edits. We introduce \ourbench, with 327 forms across Instruction-Only and Multi-File tracks. Compared with the strongest baselines, \oursys improves pair accuracy by 8.51 and 23.38 percentage points on these tracks, respectively. Component-removal experiments support structural adjudication and slot re-grounding within the pipeline, and the framework retains its relative advantage among the methods evaluated with a second backbone.
The code and benchmark \ourbench are available at {\it \textcolor{blue}{\url{https://github.com/Glllllly/CRAFT}}}.

\end{abstract}

\input{1_introduction}
\input{2_task}
\input{3_method}
\input{4_benchmark}
\input{5_experiments}
\input{6_related}
\input{7_conclusion}
\clearpage
%\section*{AI Use Statement}
%Generative AI tools were used to construct synthetic benchmark templates and instructions, as described in Section~\ref{sec:benchmark}. Claude Sonnet 4.6 was used to generate synthetic templates, and Gemini 3.1 Flash-Lite-Preview was used to generate natural-language filling instructions. LLMs and VLMs are also components of the evaluated systems, and Gemini 3.1 Flash-Lite-Preview serves as the semantic judge, as described in Section~\ref{subsec:experimental_setup}. OpenAI Codex assisted with adapting the manuscript to the ICLR LaTeX format, correcting language and typesetting issues, and condensing and reorganizing the manuscript and appendix.

\section*{AI Use Statement}
Generative AI tools were used to assist with benchmark construction, including synthetic template generation and natural-language instruction generation, as described in Section~\ref{sec:benchmark}. They were also used to support code development and improve the clarity and organization of the manuscript. LLMs and VLMs additionally serve as components of the evaluated systems and as the semantic judge, as described in Section~\ref{subsec:experimental_setup}. All AI-generated content and code were reviewed and validated by the authors, who take responsibility for the final benchmark, experiments, analyses, and claims.

\section*{Ethics Statement}
This work does not involve human subjects or the collection of personal or sensitive individual information. The benchmark is constructed from synthetic or appropriately sourced documents, with source use and redistribution checked during curation. We acknowledge that automatic form filling may introduce risks related to privacy, erroneous automated entries, and downstream use in sensitive workflows. We therefore emphasize human verification of generated outputs and document the data construction and evaluation procedures to support responsible use.

\section*{Reproducibility Statement}
To facilitate reproducibility, we provide a code repository at
{\it \textcolor{blue}{\url{https://github.com/Glllllly/CRAFT}}}
, containing the implementation of \oursys, the complete
\ourbench used in our experiments, prompts, and evaluation code.

\bibliographystyle{iclr2027_conference}
\bibliography{sample-clean}
\clearpage
\appendix
\input{appendix}
\end{document}

%% file: 1_introduction.tex
\begin{figure}[h]
    \centering
    \vspace{-1em}
    \includegraphics[width=\linewidth]{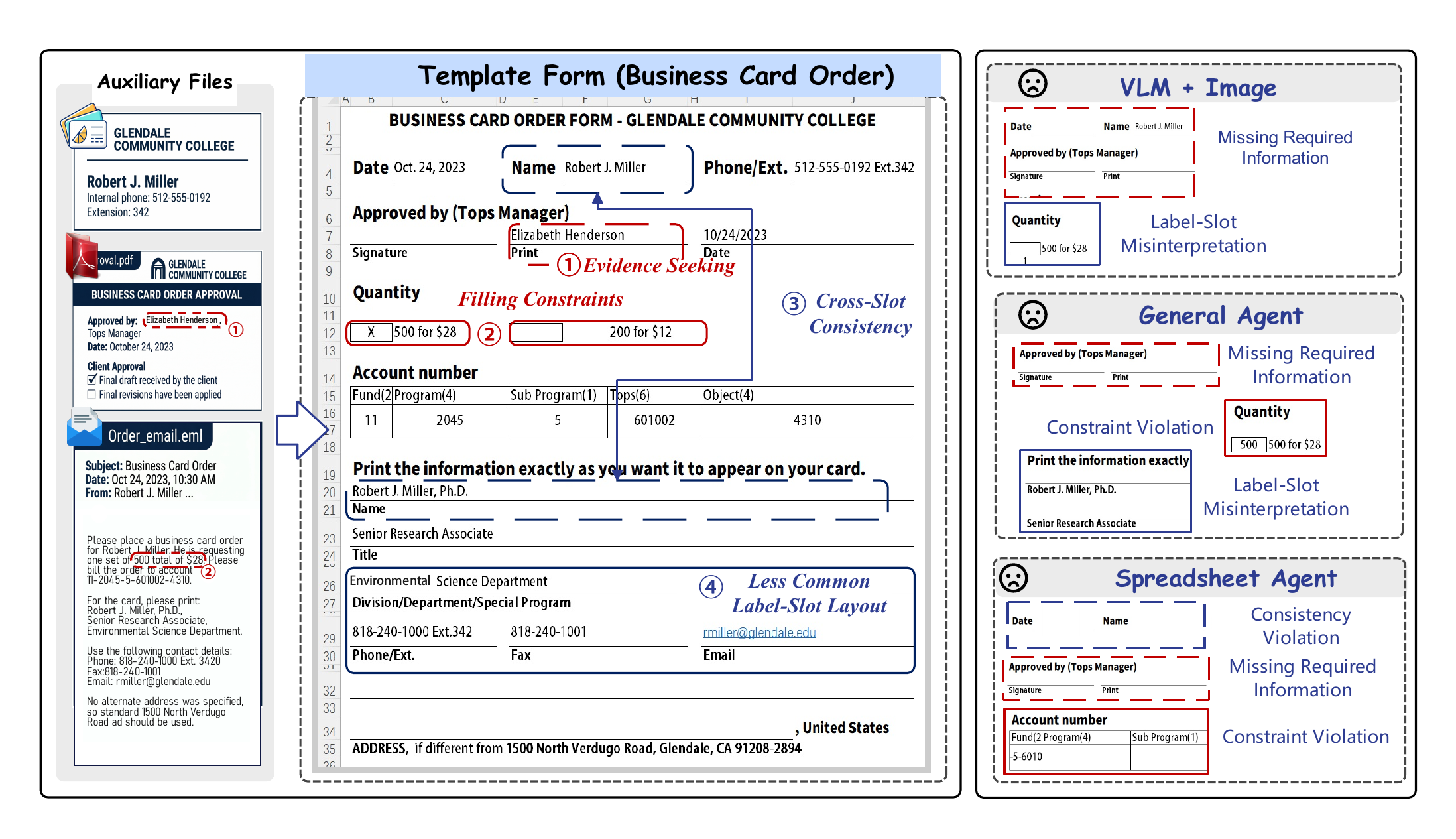}
    \vspace{-1.5em}
    \caption{\textbf{Example of spreadsheet form filling and its key challenges.} The agent needs to gather evidence from auxiliary files, satisfy field-level constraints, preserve cross-field consistency, and interpret diverse form layouts.}
    \label{fig:intro_scenarios}
\end{figure}

\section{Introduction}
\label{sec:intro}

Spreadsheet form filling requires an agent to recover information from multiple external sources and write it into a structured spreadsheet artifact without damaging its template. Unlike conventional tabular spreadsheets with regular row--column structures, forms such as reimbursement sheets and compliance questionnaires~\citep{berkeley_travel_reimbursement_xlsx,clark_vendor_cybersecurity_questionnaire_xlsx} organize information through label--slot relationships and spatial layout, where labels and writable regions may be separated, repeated, or span merged cells. Recovering the correct value is therefore insufficient: the agent must also ground it to the correct location while preserving the surrounding structure.

This challenge becomes more pronounced when the required information is scattered across heterogeneous sources.
Consider a business-card order supported by an approval PDF, an email, and an old business card (Figure~\ref{fig:intro_scenarios}). The same person appears in multiple sections, while an account number must be split across designated subfields. An agent can retrieve the right value yet write it into the wrong occurrence of a repeated label, combine subfields into one cell, or overwrite the label itself. Once such an edit changes the template, later decisions can inherit the error. Repair must therefore recover the local template context as well as the intended value.

Existing work addresses important parts of this problem. Spreadsheet agents support reasoning and manipulation~\citep{li2023sheetcopilot,chen2025sheetagent,wang2026sheetbrain,zhu2025sheetmind}, while visual form agents study interaction with web forms or image-based documents~\citep{li2025formfactory,toles2026formgym}. General agent frameworks coordinate tools and heterogeneous inputs~\citep{langgraph,openai_agents_sdk,crewai}. However, reliable spreadsheet completion requires more than recovering the correct values. The agent must also preserve the template, associate each label with the correct writable region, and maintain consistency across different parts of the form. Our pilot study finds that simply changing the relative position of a label and its corresponding slot can substantially affect the performance of existing systems, even when the information to be filled remains unchanged.

We propose \oursys, a \textbf{C}onstrained and \textbf{R}eflective \textbf{A}gent-based system for spreadsheet \textbf{F}orm filling with \textbf{T}emplate awareness. Its key idea is to couple error diagnosis with restoration and re-grounding: a suspicious region carries the evidence needed to decide what to restore, where to refill, and which cells to protect. \oursys first produces an initial filled workbook from the instruction, source evidence, and aligned structural and visual views of the template. It then converts validation signals into localized repair state: suspicious regions are identified, template corruption is restored when necessary, and plausible writable slots are re-grounded before any corrective edits are allowed. RASG provides candidate writable cells for these regions, while reliable and unaffected cells are carried forward as explicit protection constraints.

Our contributions are threefold. First, we develop a template-aware validation-and-repair framework that connects evidence consolidation, error localization, and constrained spreadsheet edits. Second, we introduce RASG, which uses rectangular slot geometry and boundary predictions to guide local repair under irregular layouts. Third, we construct \ourbench, comprising 327 forms across Instruction-Only and Multi-File tracks, and evaluate \oursys against 11 baselines. \oursys improves pair accuracy over the strongest baseline by 8.51 and 23.38 percentage points on the two tracks, respectively. Component-removal experiments assess structural adjudication, slot re-grounding, and planning within the full pipeline, and an additional backbone evaluation tests the framework beyond the default model.

%% file: 2_task.tex
\section{Task and Motivation}
\label{sec:preliminaries}
\label{sec:form_filling_task}

\paragraph{Spreadsheet form filling.}
Given a template form $\mathcal{F}$, an instruction $\mathcal{X}$, and auxiliary files $\mathcal{S}$, the task is to produce filled entries $\mathcal{O}=\{(\ell_i,v_i,R_i)\}_{i=1}^{N}$. Here $\ell_i$ identifies a label, $v_i$ is its value, and $R_i$ is the target cell or range. A label identifier combines its text with an occurrence index to distinguish repeated labels. Unsupported values are assigned $\text{null}$. Applying these entries should fill the intended slots while preserving static labels, formatting, merged regions, and other template constraints. Appendix~\ref{app:formalism} gives the full form representation and examples.

% Predicted and ground-truth entries are aligned by their composite label identifiers.

Predicted and ground-truth entries are aligned using their composite label identifiers. For each aligned pair, correctness requires both value agreement and exact spatial grounding:
\begin{equation}
\operatorname{Correct}(e_i,e_i^*)
=
\mathbb{I}\!\left[
\operatorname{MatchValue}(v_i,v_i^*)
\land
(R_i=R_i^*)
\right].
\end{equation}
Value matching accepts semantically equivalent strings, such as alternative date formats. Values spanning multiple cells or lines are concatenated before comparison. This separates semantic equivalence from the stricter requirement that values occupy the correct spreadsheet range.

\begin{figure}[!t]
    \centering
    \includegraphics[width=0.8\linewidth]{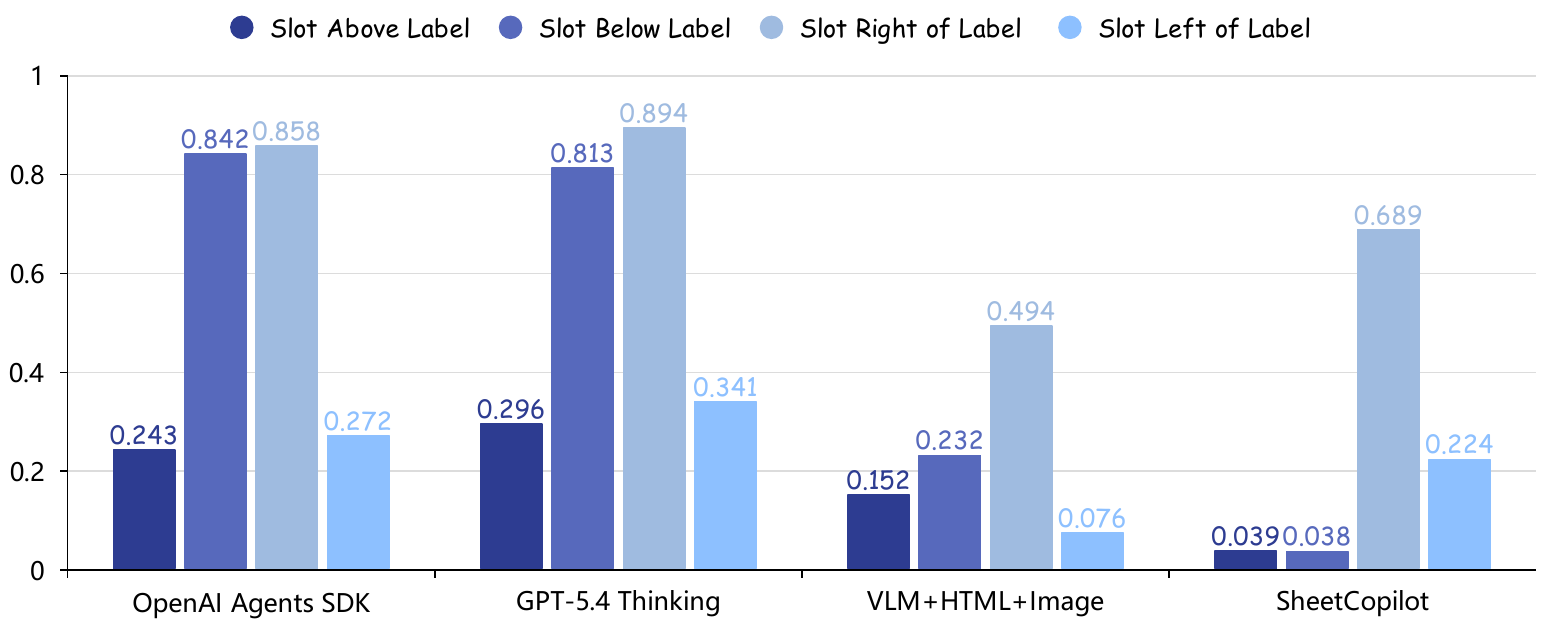}
    \caption{Pilot Study of Spreadsheet Form Filling. Samples are from two common categories of real spreadsheet forms (i.e., mixed label-slot layout and repeated-section forms).} 
    \label{fig:pilot-study}
\end{figure}

\paragraph{Why localized repair?}
\label{sec:pilot-study}
We conduct a pilot study to investigate how sensitive existing systems are to variations in label--slot layouts. We randomly select 15 mixed-layout and repeated-section forms and expand them into 57 pilot instances by varying label--slot positions while keeping filling semantics fixed. For this diagnostic study, we additionally include GPT-5.4 Thinking~\citep{openai2026gpt54} to examine layout sensitivity under a reasoning-oriented model, complementary to the lightweight backbones adopted in the main evaluation for large-scale experimentation under a practical inference budget. Figure~\ref{fig:pilot-study} shows that performance depends on label--slot direction; the left-label/right-slot arrangement is generally favorable, although sensitivity varies across systems. Failures include writing into labels, selecting neighboring cells on the wrong side, and producing inconsistent entries across repeated sections; error locations also vary across runs. The full study appears in Appendix~\ref{app:pilot}. These observations motivate validation against the original template and explicit re-grounding of suspicious regions before further edits.

%% file: 3_method.tex
\section{CRAFT}

\begin{figure}[t]
\centering
\includegraphics[width=\linewidth]{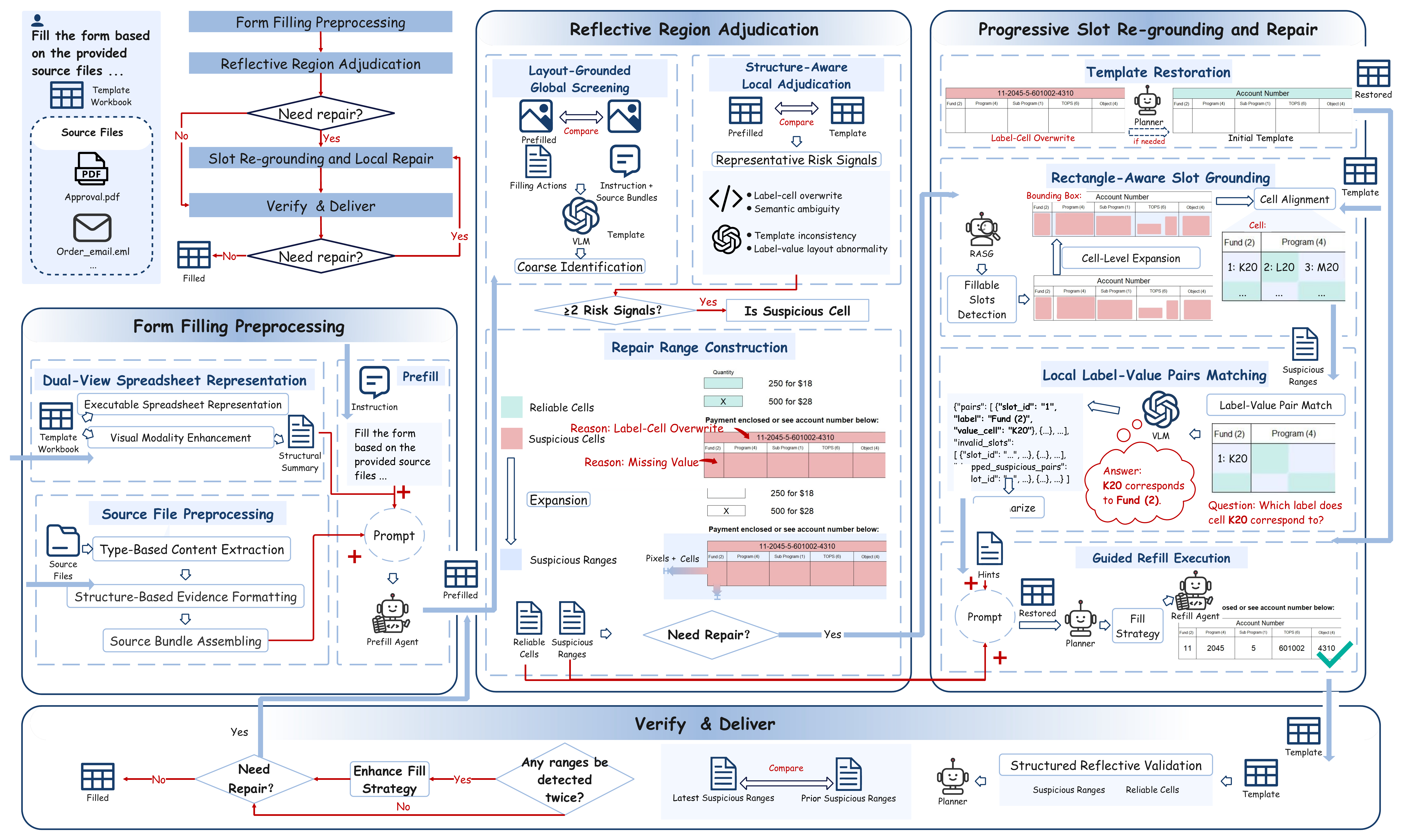}
\caption{\oursys combines evidence-based prefilling, reflective region adjudication, and constrained local repair. The planner carries validation evidence and protected cells between stages.}
\label{fig:framework}
\end{figure}

\hi{Architecture.}
Figure~\ref{fig:framework} illustrates the overall architecture of \oursys, which consists of three modules coordinated by an agent-based planner. \bfit{(1) Form Filling Preprocessing} normalizes heterogeneous auxiliary files into structured evidence and constructs a dual-view spreadsheet representation with an executable view for precise manipulation and a layout-aligned visual view for VLM-based reflection (Section~\ref{sec:preprocessing}). \bfit{(2) Reflective Region Adjudication} combines global screening and structure-aware local validation to identify reliable and suspicious cells, which are consolidated into suspicious ranges for repair (Section~\ref{sec:srv}). \bfit{(3) Progressive Slot Re-grounding and Repair} restores corrupted template content, re-identifies plausible writable slots, matches them to labels, and performs constrained local refill (Section~\ref{sec:guided_refill}).

\hi{Workflow.}
Given a template form, user instruction, and auxiliary files, \oursys first preprocesses the inputs and produces an initial prefilled form. The planner applies reflective adjudication to localize suspicious ranges and progressive repair to restore template structure and refill the affected slots. Execution is budgeted: the evaluated configuration permits at most one reflection round and six patch steps after prefilling (Section~\ref{subsec:experimental_setup}). The returned workbook may therefore retain unresolved errors.

%\hi{Running Example.} For clarity, we use the business-card scenario in Figure~\ref{fig:intro_scenarios} as a running example throughout.

\subsection{Form Filling Preprocessing}
\label{sec:preprocessing}

%\yeye{if running out of space, I would recommend to shorten Section 5-8, highlight key points from each step and keeping them concise. Right now these sections are a little ``dry'' to read, would be better to connect back to the example in Figure 1, to make them intuitive / easy-to-follow (There is an example 3, which is good).}

Evidence for form filling is often scattered across heterogeneous sources, such as emails, PDFs, and spreadsheets, whose structural and spatial cues can be lost when flattened into one-dimensional sequences~\citep{xu2020layoutlm, Li0W0025}. Our preprocessing converts these sources into a unified, structure-aware evidence representation for downstream form filling.

\paragraph{Dual-View Spreadsheet Representation.}
\label{subsec:preprocessing_workspace}
\oursys first parses the spreadsheet template to recover cell coordinates, and determines the exact form range from the maximal populated row--column extent. Then, \oursys derives two complementary views over this range. The \emph{executable view} retains cell coordinates, content, formatting, merged regions, writable and non-writable areas, and constraints such as dropdown candidate values. The \emph{layout view} renders structured markup into a sheet image and records its alignment with spreadsheet coordinates. The first view supports precise edits and programmatic checks; the second exposes spatial relationships for VLM-based validation. The alignment between the two views bridges visual understanding and spreadsheet manipulation. It allows the system to ground VLM-identified anomalies to concrete cells or ranges for subsequent repair.

\paragraph{Source File Preprocessing.}
\label{subsec:preprocessing_sources}
\oursys performs \emph{Type-Based Content Extraction}, \emph{Structure-Based Evidence Formatting}, and \emph{Source Bundle Assembling}. Extraction preserves source-specific cues: text spans and sections, structured label hierarchies, table headers and records, email headers and body sections, and PDF text blocks, tables, and layout anchors. Lightweight signals---label--value regularity, section patterns, and repeated records---then guide formatting into label--value pairs, sectioned text, tables, hierarchical objects, or hybrid views. Weakly structured sources retain grounded text spans. The resulting source bundle provides structured evidence for the planner.

\paragraph{Planner-guided prefilling.}
\label{subsec:preprocessing_prefill}
The planner combines the instruction, template summary, and source bundle to associate target regions with relevant evidence and establish filling policies, including how to handle unsupported fields. A prefill agent executes this plan through spreadsheet tools. When the workbook is too large to inline, the agent instead inspects its structure and reads small local ranges relevant to each filling decision. The resulting workbook is an initial candidate for validation, not an assumption of correct completion.

\begin{example}[Preprocessing]
In the running case (Figure~\ref{fig:framework}), an approval PDF, an email, and a business-card image supply the evidence. Signals such as \texttt{kv\_line\_ratio} and \texttt{section\_line\_ratio} guide formatting into sectioned blocks, including ``root'' and ``Use the following contact details.'' The planner combines these source bundles with the instruction and aligned template views. It assigns A8:T11 to approval information from Approval.pdf and records that the account number must be split across designated fields.
\end{example}

\subsection{Reflective Region Adjudication}
\label{sec:srv}

The prefilled form may contain plausible but incorrect filling actions, such as overwriting labels, placing values into adjacent cells, or violating form-specific constraints. These errors are difficult to detect in isolation and can substantially degrade downstream accuracy~\citep{stechly2024self,huang2023large}. To address this issue, \oursys performs (1) local reflective, evidence-grounded validation to identify reliable and suspicious cells, and (2) range construction to consolidate suspicious cells into compact regions for downstream repair.

\paragraph{Global screening and local adjudication.}
In the first stage, a VLM compares aligned views of the original and prefilled workbooks, conditioned on the instruction and edit history. It identifies reliable cells and suspicious regions, including missing values, abnormal overwrites, and inconsistent placement. In the second stage, Structure-Aware Local Adjudication (SALA) diagnoses candidates using spreadsheet-specific signals. Programmatic checks detect overwritten static template cells; VLM judgments assess label--value ordering, separation, and other layout-grounding risks. The local stage marks a cell suspicious when at least two risk signals fire, reducing repairs triggered by a single noisy cue. This threshold applies to local adjudication; a coarse warning from global screening can also contribute a repair candidate.

\paragraph{Repair range construction.}
\label{subsec:srv_ranges}
The system merges the two stages' judgments and expands suspicious cells into coherent local ranges. Expansion first occurs in spreadsheet space, treating merged cells as atomic units, then in pixel space to retain borders and nearby layout context. Each range carries its error types, validation provenance, restoration indicator, and reliable cells. A template-integrity violation triggers restoration before refill, whereas a layout or content anomaly may require only a local rewrite. The planner uses these outputs to decide which ranges enter repair. The representation preserves the context needed for grounding while protecting reliable content outside, and within, the selected region.

 \begin{example}[Reflective Region Adjudication]
For the running case in Figure~\ref{fig:framework}, the Reflective Region Adjudication round produces several suspicious cells (or merged-cell regions). Concretely, \texttt{K19:T19} is identified by the Structure-Aware Local Adjudication stage because it triggers both \textit{label-cell overwrite} and \textit{label--value layout abnormality}, while \texttt{K21:T22} is identified by the Layout-Grounded Global Screening stage due to an explicit missing value. These suspicious cells are first unionized and expanded in spreadsheet space to form a compact cell-level repair range, \texttt{G18:T23}. This cell range is then mapped to its corresponding rendered region in the layout view, where an additional pixel-level margin is added to preserve surrounding layout and semantic context for subsequent repair.
\end{example}

\subsection{Progressive Slot Re-grounding and Repair}
\label{sec:guided_refill}

\paragraph{Template restoration.}
Within a range flagged for template-integrity violations, \oursys compares cell values, styles, and merged-range structure against an immutable template snapshot. It restores overwritten template content while preserving cells already judged reliable. This recovers the local structure on which subsequent grounding depends.

\begin{figure}[t]
\centering
\includegraphics[width=\linewidth]{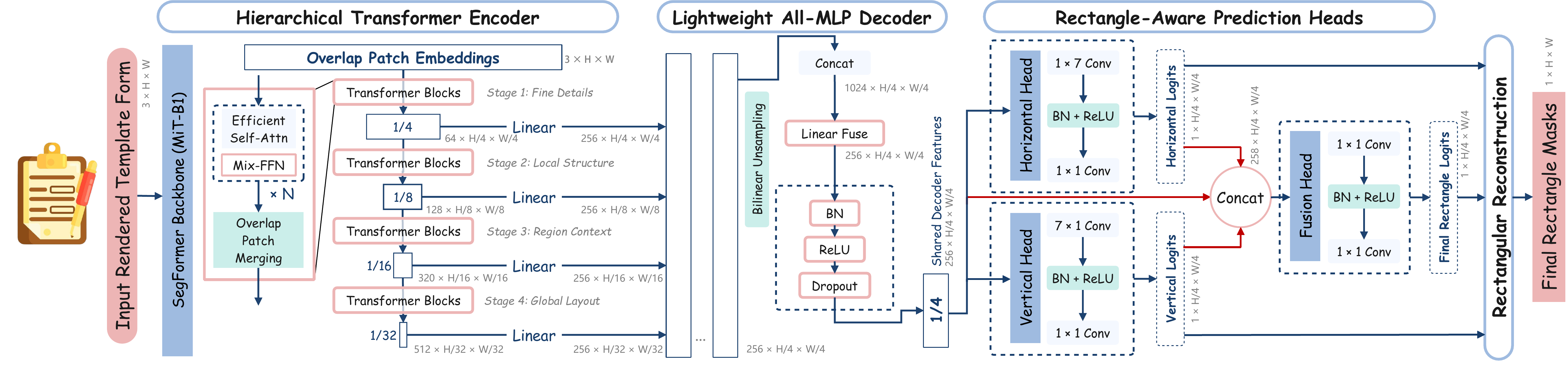}
\caption{Rectangle-Aware Slot Grounder (RASG). A SegFormer-B1 backbone supports rectangle prediction and auxiliary horizontal and vertical boundary heads.}
\label{fig:RASG}
\end{figure}

\paragraph{Rectangle-aware slot grounding.}
RASG localizes plausible writable cells in the restored layout view (Figure~\ref{fig:RASG}). A SegFormer-B1 backbone~\citep{xie2021segformer} encodes four scales of features, which are projected, upsampled, and concatenated into a shared decoder representation. Three heads predict rectangle logits and horizontal and vertical boundary cues. Boundary predictions rectify and reconstruct rectangle masks, which are aligned with candidate writable cells using area-based thresholds. These candidates narrow the locations considered by the refill agent.
We provide qualitative visualizations of the directional boundary cues and reconstructed slot predictions in Appendix~\ref{app:rasg_visualization}.

RASG is trained in three stages: synthetic spreadsheet forms provide bounding-box and cell-level supervision; CommonForms~\citep{barrow2026commonforms} supplies more varied PDF layouts with bounding-box supervision; and a manually filtered, annotated FUSE subset~\citep{barik2015fuse} provides adaptation to real spreadsheet forms with both supervision types. This progression introduces more realistic layouts while retaining supervision for spreadsheet slot localization. If no reliable slot is found, the controller falls back to guarded local repair using the suspicious range, template summary, and protection policy.

\paragraph{Label matching and constrained refill.}
A VLM checks each candidate value cell and associates valid candidates with nearby labels. Valid matches become positive refill hints; rejected candidates become negative hints excluding non-target locations. Context-only hints retain useful label information when a target cannot yet be selected. The refill agent combines these hints with the instruction, restored workbook state, and validation feedback to generate a local plan specifying allowed edits and protected cells. It then applies spreadsheet actions within this specification. Thus, reflection is connected to explicit spatial targets and preservation constraints rather than a free-form request to regenerate the form.

\begin{example}[Progressive Slot Re-grounding and Local Repair]
For the same running case in Figure~\ref{fig:framework}, reflective validation identifies the suspicious range \texttt{G18:T23}. Since this range triggers the \texttt{label-cell overwrite} signal, \oursys first applies template restoration to recover the original template content, including the overwritten label Account Number. The restored local region is then passed to the RASG, which detects a set of plausible writable slots and then aligns them with candidate cells such as \texttt{K20}, \texttt{L20}, and \texttt{M20}. Based on these candidates and the rendered local context, a VLM further performs local label--value matching and determines, for example, that cell \texttt{K20} corresponds to the label ``\texttt{Fund (2)}''. These outputs are summarized into structured refill hints, which guide the refill agent to perform constrained local rewriting within the suspicious range and correctly fill the segmented account-number fields.
\end{example}

%% file: 4_benchmark.tex
\section{FormFillBench}
\label{sec:benchmark}

Despite rapid progress in LLM-based agents for document interaction and spreadsheet automation~\citep{lin2025querying, chen2025sheetagent}, existing form filling benchmarks mainly target web-based or document-centered settings~\citep{li2025formfactory, toles2026formgym}, while existing spreadsheet evaluation efforts focus on spreadsheet understanding and manipulation rather than spreadsheet form filling~\citep{rahman2020benchmarking}. To our knowledge, spreadsheet form filling has not been explicitly studied as a dedicated benchmark task. To address this gap, we introduce \ourbench.

\ourbench contains 327 form instances across two tracks. The \emph{Instruction-Only Track} has 200 forms and 5,499 label--value pairs; all filling information is supplied in a natural-language instruction. The \emph{Multi-File Track} has 127 forms and 2,288 pairs, with roughly five supporting files per case and 11 file types overall. It tests evidence consolidation in addition to spatial grounding.

Templates come from FUSE~\citep{barik2015fuse}, VEUSES and Venron2~\citep{xu2017spreadcluster,dou2016venron}, public online sources, and synthetic templates generated with Claude Sonnet 4.6~\citep{anthropic2026sonnet46}. For the Instruction-Only Track, Gemini 3.1 Flash-Lite-Preview generates instructions from rendered forms and key annotations. Multi-File cases combine open-source resources and synthetic enterprise documents under a standardized instruction to use explicitly supported information. Candidate samples undergo automated consistency checks and manual verification. Appendix~\ref{app:benchmark} details construction, file distributions, and five form archetypes spanning regular label--slot layouts, mixed layouts, repeated sections, mixed sections, and bordered blocks.

The two tracks separate complementary difficulties. Instruction-Only cases expose whether a system can ground supplied values into the right locations; Multi-File cases additionally require source selection and evidence consolidation. Structural diversity occurs within both tracks: regular left--right layouts coexist with mixed label--slot directions, repeated groups, and sections with different internal organizations. The archetypes summarize these patterns rather than constitute a strict taxonomy. This makes it possible to inspect where performance varies instead of relying solely on an aggregate accuracy.

\paragraph{Metrics.}
\emph{Pair accuracy} measures the proportion of target pairs whose values and ranges are both correct. \emph{End-to-end (E2E) accuracy} measures the fraction of forms with every target pair correct, including fields that should remain blank. These primary metrics apply to both tracks and measure target-entry correctness, rather than a separate audit of every formatting or template-preservation constraint. Supplementary metrics distinguish filling from leaving fields blank: fill precision, recall, and F1 for Instruction-Only, and fill coverage (reported as Fill Accuracy) for Multi-File. Their definitions and complete scores are given in Appendix~\ref{app:benchmark} and Table~\ref{tab:full_results}.

%% file: 5_experiments.tex
\section{Experiments}
\label{sec:experiments}

\subsection{Experimental Setup}
\label{subsec:experimental_setup}

We compare \oursys with 11 baselines: three direct VLM variants (HTML, HTML plus image, and coding with basic Excel skills); four general agent methods (LangGraph, OpenAI Agents SDK, CrewAI, and ReAct); and four spreadsheet agents (SheetAgent, SheetBrain, SheetCopilot, and SheetMind). Section~\ref{sec:relatedwork} discusses these methods. Target spreadsheets are normalized to HTML and non-spreadsheet sources to Markdown. The HTML-plus-image baseline additionally receives a rendered sheet image. Tool-based methods share a spreadsheet interface, while direct-prompting methods operate on their provided renderings.

Unless otherwise specified, \oursys and agent-based baselines use Gemini 3.1 Flash-Lite-Preview~\citep{google2026gemini31flashlite} with temperature 1.0. The same model acts as the semantic-only judge. \oursys uses structure-scanning, range-reading, and cell-writing tools, with up to 10 main interaction steps, at most one reflection round, and up to 6 patch steps. Agent-based baselines use the same temperature and interaction budget. This protocol aligns source information and interaction budgets while retaining differences in the methods' native interfaces; it does not equate token usage or monetary cost.

\input{tables/main_results}

\subsection{Main Results}
\label{subsec:main_results}

Table~\ref{tab:main_results} reports the two primary metrics. \oursys leads on both, reaching 59.68 pair accuracy and 15.5 E2E accuracy on \emph{Instruction-Only}, and 66.52 and 43.31 on \emph{Multi-File}. Relative to VLM + HTML + Image, the strongest Instruction-Only baseline, the gains are 8.51 and 2.00 percentage points. On Multi-File, the best baseline differs by metric: OpenAI Agents SDK~\citep{openai_agents_sdk} reaches 43.14 pair accuracy, while CrewAI reaches 18.90 E2E accuracy. The corresponding gains are 23.38 and 24.41 points.

The baseline rankings are consistent with different task bottlenecks. In the \emph{Instruction-Only Track}, strong VLMs can recover many local label--value correspondences when the form layout is clearly exposed, but remain vulnerable to structural execution errors such as writing to nearby cells, misaligning values with slots, or overwriting template content. In the \emph{Multi-File Track}, the challenge shifts upstream to evidence aggregation across heterogeneous sources, label alignment, and consistent multi-step execution, making general agent frameworks relatively more competitive. Existing spreadsheet agents still underperform both strong VLM baselines and general agents, likely because they are optimized for spreadsheet manipulation rather than fine-grained grounding and constraint-sensitive form filling. The tracks contain different form sets, so their absolute scores are not a controlled comparison of task difficulty.

% \oursys reaches 59.68 pair accuracy and 15.5 E2E accuracy on Instruction-Only, compared with 51.17 and 13.5 for the strongest direct VLM baseline. On Multi-File, it reaches 66.52 pair accuracy and 43.31 E2E accuracy. The strongest baseline pair accuracy is 43.14 (OpenAI Agents SDK), while the highest baseline E2E accuracy is 18.90 (CrewAI). The gains are therefore 8.51 and 23.38 percentage points in pair accuracy, and 2.0 and 24.41 points in E2E accuracy, across the respective tracks.

% The strongest baseline changes with the track. Direct prompting is competitive when the instruction already contains the necessary values and a rendering exposes the form layout. With heterogeneous auxiliary files, general agent frameworks become more competitive, consistent with the additional need to gather and consolidate evidence. Spreadsheet agents lag behind both groups, with SheetCopilot the strongest among them. Their focus on manipulation of existing spreadsheets appears insufficient for the combined evidence-grounding and template-preservation requirements of this task.

\input{tables/ablation}

\subsection{Ablation and Backbone Transfer}
\label{subsec:ablation}

Table~\ref{tab:ablation_main} compares four component-removal variants. Removing SALA reduces pair accuracy by 6.71 points on \emph{Instruction-Only} and 41.04 points on \emph{Multi-File}. Removing Progressive Slot Re-grounding (PSR) reduces it by 9.23 and 41.66 points, respectively. The latter variant also lowers \emph{Multi-File} E2E accuracy from 43.31 to 3.15. These results support structural adjudication and slot re-grounding as complementary parts of the evaluated pipeline.

Removing the entire reflection stage yields 53.71 and 24.96 pair accuracy across the two tracks. On Instruction-Only, its 3.26-point advantage over the variant without PSR (50.45) is consistent with repair becoming counterproductive without re-grounding. The corresponding Multi-File scores, 24.96 and 24.86, are nearly tied. Removing both reflection and planner-controlled execution yields 51.63 and 24.17. The non-monotonic E2E scores on \emph{Instruction-Only} further indicate that the effects of individual components depend on the metric and setting; the full system provides the strongest overall results.

\input{tables/backbone}

With GPT-5.4 mini~\citep{openai2026gpt54mininano} as the backbone, \oursys maintains the lead among the evaluated methods (Table~\ref{tab:change model}). It reaches 41.57 pair accuracy on \emph{Instruction-Only} and 60.66 on \emph{Multi-File}, exceeding the strongest corresponding baselines by 7.51 and 10.75 points. This supports the relative benefit of the framework with a second backbone. Absolute performance remains model-dependent: Multi-File E2E accuracy is 9.45 with GPT-5.4 mini, compared with 43.31 with the default backbone.

\subsection{Case Study}

We further examine representative cases from both the \bfit{Instruction-Only Track} and the \bfit{Multi-File Track}. Three recurring patterns emerge. \textit{First}, vanilla VLM baselines remain competitive on visually regular layouts. For example, on the Left-Right Label-Slot case (Case 38), multiple rendering-based baselines achieve full success. However, their performance degrades on more irregular structures, such as the Mixed-Slot-Layout case (Case 29), where vanilla VLM baselines achieve only partial success. \textit{Second}, spreadsheet-oriented agents show limited robustness across structurally diverse forms. Although \textsc{SheetCopilot}~\citep{li2023sheetcopilot} succeeds on the Mixed-Section case (Case 88) and partially succeeds on the Left-Right Label-Slot case (Case 38), spreadsheet agents fail on most other cases, including the Repeated-Section case (Case 4), the Mixed-Slot-Layout case (Case 29), and the Bordered-Block case (Case 84). \textit{Third}, the \emph{Multi-File Track} remains challenging for nearly all baselines. On the Bordered-Block (Case 226), Mixed-Slot-Layout (Case 238), and Repeated-Section (Case 284) cases, most competing methods achieve only partial success or failure, indicating difficulties in consolidating evidence across files and grounding retrieved values back to the target form. Across all selected cases, \oursys consistently achieves successful completion.
% We further examine representative examples from both the \emph{Instruction-Only Track} and the \emph{Multi-File Track} to analyze common success and failure patterns across methods. Three recurring observations emerge. \textit{First}, vanilla VLM baselines remain competitive on visually regular single-form layouts, but become less reliable when the form contains irregular slot arrangements or ambiguous local label--value relationships. \textit{Second}, spreadsheet-oriented agents show limited robustness beyond relatively simple spreadsheet-native settings, frequently failing on structurally diverse forms that require fine-grained slot grounding. \textit{Third}, the \emph{Multi-File Track} remains challenging for nearly all baselines, particularly when values must be consolidated across heterogeneous sources and grounded back to the correct target slots. Across these representative examples, \oursys shows substantially more consistent completion behavior.
Detailed case-level results and representative examples are provided in Appendix~\ref{sec:case_study}.

%% file: tables/main_results.tex
\begin{table}[t]
\centering
\caption{Primary benchmark results (\%). Pair accuracy requires both the value and target range to be correct; E2E requires every pair in a form to be correct. All 11 baselines are shown. Complete metrics appear in Table~\ref{tab:full_results}.}
\label{tab:main_results}
\small
\begin{tabular}{@{}lrrrr@{}}
\toprule
& \multicolumn{2}{c}{\textbf{Instruction-Only}} & \multicolumn{2}{c}{\textbf{Multi-File}} \\
\cmidrule(lr){2-3}\cmidrule(lr){4-5}
\textbf{Method} & \textbf{Pair} & \textbf{E2E} & \textbf{Pair} & \textbf{E2E} \\
\midrule
VLM+HTML+Image & 51.17 & 13.5 & 32.60 & 12.60 \\
VLM+HTML & 49.26 & 11.0 & 26.31 & 7.87 \\
VLM+Excel Skills+Coding & 10.66 & 0.0 & 0.26 & 0.0 \\
\midrule
LangGraph & 44.28 & 10.0 & 33.65 & 12.60 \\
OpenAI Agents SDK & 45.64 & 9.5 & 43.14 & 16.54 \\
CrewAI & 41.77 & 9.5 & 40.25 & 18.90 \\
ReAct & 41.66 & 11.0 & 34.00 & 13.39 \\
\midrule
SheetAgent & 8.89 & 0.0 & 0.26 & 0.0 \\
SheetBrain & 10.98 & 0.0 & 0.26 & 0.0 \\
SheetCopilot & 33.02 & 7.5 & 0.26 & 0.0 \\
SheetMind & 10.69 & 0.0 & 0.26 & 0.0 \\
\midrule
\oursys & \textbf{59.68} & \textbf{15.5} & \textbf{66.52} & \textbf{43.31} \\
\bottomrule
\end{tabular}
\end{table}

%% file: tables/ablation.tex
\begin{table}[t]
    \centering
    \caption{Ablation results (\%). SALA: Structure-Aware Local Adjudication, PSR: Progressive Slot Re-grounding, Re.: Whole Reflection module, Pl.: planner.}
    \label{tab:ablation_main}
    \small
    \setlength{\tabcolsep}{3pt}
    \begin{tabular}{lcc|cc}
    \toprule
    & \multicolumn{2}{c|}{\textbf{Instruction-Only}} & \multicolumn{2}{c}{\textbf{Multi-File}} \\
    \cmidrule(lr){2-3} \cmidrule(lr){4-5}
    \textbf{Method} & \textbf{Pair Acc.$\uparrow$} & \textbf{E2E Acc.$\uparrow$} & \textbf{Pair Acc.$\uparrow$} & \textbf{E2E Acc.$\uparrow$} \\
    \midrule
    \oursys & \textbf{59.68} & \textbf{15.5} & \textbf{66.52} & \textbf{43.31} \\
    \oursys w/o SALA & 52.97 & 10.0 & 25.48 & 10.24 \\
    \oursys w/o PSR & 50.45 & 12.5 & 24.86 & 3.15 \\
    \oursys w/o Re. & 53.71 & 10.5 & 24.96 & 7.87 \\
    \oursys w/o Re. \& Pl. & 51.63 & 12.5 & 24.17 & 6.30 \\
    \bottomrule
    \end{tabular}
\end{table}

%% file: tables/backbone.tex
\begin{table}[t]
    \centering
    \caption{Results with GPT-5.4 mini (\%).}
    \label{tab:change model}
    \small
    \setlength{\tabcolsep}{3pt}
    \begin{adjustbox}{max width=\columnwidth}
    \begin{tabular}{lcc|cc}
    \toprule
    & \multicolumn{2}{c|}{\textbf{Instruction-Only}} & \multicolumn{2}{c}{\textbf{Multi-File}} \\
    \cmidrule(lr){2-3} \cmidrule(lr){4-5}
    \textbf{Method} & \textbf{Pair Acc.$\uparrow$} & \textbf{E2E Acc.$\uparrow$} & \textbf{Pair Acc.$\uparrow$} & \textbf{E2E Acc.$\uparrow$} \\
    \midrule
    \oursys & \textbf{41.57} & \textbf{9.5} & \textbf{60.66} & \textbf{9.45} \\
    VLM + HTML + Image & 34.06 & 8.5 & 25.86 & 5.95 \\
    OpenAI Agents SDK  & 29.24 & 2.0 & 49.91 & 3.54 \\
    SheetCopilot & 12.75 & 0.5 & 0.66 & 0.0 \\
    \bottomrule
    \end{tabular}
    \end{adjustbox}
\end{table}

%% file: 6_related.tex
\section{Related Work}
\label{sec:relatedwork}

\hi{Form and Document Understanding.}
Recent work studies form interaction and template-aware document processing through multimodal perception, including visual form agents~\citep{toles2026formgym,li2025formfactory}, structured information extraction~\citep{majumder-etal-2020-representation,cao2021extracting}, and template-based reconstruction or querying~\citep{lin2025visual,lin2025twix,lin2024zendb,lin2025querying}. FormGym's FieldFinder~\citep{toles2026formgym} and CommonForms~\citep{barrow2026commonforms} make field localization a relevant point of comparison. \oursys couples slot localization to spreadsheet repair: it restores overwritten template content, maps candidates to executable cell ranges, and protects reliable cells during refill.

\hi{Spreadsheet Agents.}
Spreadsheet agents such as SheetCopilot~\citep{li2023sheetcopilot}, SheetAgent~\citep{chen2025sheetagent}, and SheetBrain~\citep{wang2026sheetbrain} support spreadsheet manipulation, planning, and validation. Reflection and validation are already part of this literature: SheetAgent uses iterative reflection, and SheetBrain validates execution results. Our contribution is the template-specific connection between diagnosis and repair: suspicious ranges carry restoration decisions and protection constraints into slot re-grounding and local refill.

\hi{General-Purpose LLM Agents.}
General frameworks such as LangGraph~\citep{langgraph}, OpenAI Agents SDK~\citep{openai_agents_sdk}, CrewAI~\citep{crewai}, and ReAct~\citep{yao2022react} enable multi-step reasoning and tool use, while Reflexion~\citep{shinn2023reflexion}, Self-Refine~\citep{madaan2023self}, and CRITIC~\citep{gou2023critic} study iterative refinement. \oursys instantiates refinement with spreadsheet-specific spatial targets, template restoration, and edit-protection constraints.

%% file: 7_conclusion.tex
\section{Conclusion}
\label{sec:conclusion}

Automatic spreadsheet form filling is essential in real-world enterprise settings. In this paper, we introduced an agent-based spreadsheet form filling system \oursys. We developed \emph{Form Filling Preprocessing} that transforms diverse multi-modal auxiliary sources into structured evidence via a dual-view spreadsheet representation. We proposed \emph{Reflective Region Adjudication} that automatically detects and localizes structural violations and content inconsistencies to identify suspicious edits under template constraints. We proposed \emph{Progressive Slot Re-grounding and Repair} that enhances filling accuracy by constraining agents to plausible writable slots while protecting reliable cells. Experimental results on \ourbench demonstrated the effectiveness of our approach over baseline methods.

\oursys is limited by the available source parsers and may require human clarification for ambiguous labels or multiple specifications within a cell. The benchmark is curated, and the reported point estimates cover two backbones and an LLM-based semantic judge. They establish performance under this evaluation protocol, not reliability across arbitrary enterprise workflows. Strict form-level completion remains challenging, especially on Instruction-Only forms and with the second backbone. Broadening evaluation coverage and quantifying run-to-run variability are directions for future work.

%% file: appendix.tex
%查有无和正文重复的部分，如有，则调整

\section{Task Formalism}
\label{app:formalism}

In this section, we first introduce the elements of spreadsheet forms. Then we give the definition of spreadsheet form filling. 

\subsection{Spreadsheet Forms}
\label{app:subsec_form_elements}

\hi{Form Elements.} We first define the core elements that determine how information is organized in a spreadsheet form and where values should be written. 

\noindent$\bullet$ \bfit{Form Range.} A spreadsheet form is inherently structurally grounded. A \textit{range} $R$ is an explicit extent in a spreadsheet form, formally represented as a set of contiguous units $R = \{u_1, u_2, \dots, u_m\}$. In our formulation, each unit $u_j$ is a \textit{cell} $c = (pos, val, fmt)$ consisting of a position $pos$ (e.g., A1), a value $val$ (e.g., text, number), and formatting attributes $fmt$ (e.g., color, font, border style).

\noindent$\bullet$ \bfit{Form Label.} Following the common paired formulation used in document understanding~\citep{xu2020layoutlm} and key information extraction~\citep{yu2021pick}, a \textit{label} represents the semantic descriptor of a fillable target in a spreadsheet form, such as ``Name'' or ``Department''. To distinguish repeated labels (e.g., multiple ``Name'' entries in duplicated sections), we define a composite \textit{label identifier} $\ell = (\text{label}, \text{label\_num})$, where $\text{label}$ is the label text and $\text{label\_num}$ indexes repeated occurrences of the same label. If a label appears only once, $\text{label\_num} = 0$.

\noindent$\bullet$ \bfit{Label--Value Pair.}
A \emph{label--value pair} is the fundamental semantic--structural unit in a spreadsheet form. Unlike JSON or relational tables, the correspondence between a label and its writable range in a spreadsheet form is often implied through layout cues such as alignment and proximity. Specifically, a label--value pair is defined as $p = (\ell, R, \tau)$, where $\ell$ is the label identifier, $R$ denotes the associated writable range for filling, and $\tau \in \mathcal{T}$ represents the local structural relation between the label and its target range (e.g., left--right alignment or top--down alignment).

\begin{definition}[Spreadsheet Form]
A spreadsheet form $\mathcal{F}$ is a layout-rich spreadsheet object designed to organize and collect structured information. It is formally defined as a tuple $\mathcal{F} = (\mathcal{P}, \mathcal{C}, \mathcal{T})$, where:
\begin{itemize}[leftmargin=*]
    \item $\mathcal{P}$ is the set of \textit{label--value pairs} that define the core fillable targets.
    \item $\mathcal{C}$ is the set of \textit{auxiliary semantic contents} (e.g., section titles, instruction texts, and notes). These elements do not correspond to fillable targets, but provide important semantic and structural cues for interpreting the spreadsheet form.
    \item $\mathcal{T}$ captures the positional and structural relations among elements, including relative positions, grouping relations, hierarchical structures, and form-level layout organizations across spreadsheet regions.
    
\end{itemize}
\end{definition}

\subsection{Form Filling Task}
\label{app:sec_form_filling_task}

Given a template spreadsheet form $\mathcal{F}$, an input text $\mathcal{X}$ (typically a natural language description), and a set of auxiliary source files $\mathcal{S}$, the spreadsheet form filling task aims to accurately extract and consolidate relevant information from $\mathcal{X}$ and $\mathcal{S}$, then write it into the corresponding spatial regions in $\mathcal{F}$.

\begin{definition}[Spreadsheet Form Filling]
\label{app:def_form_filling}
The spreadsheet form filling task aims to produce a set of filled entries $\mathcal{O}$ that specifies what value should be written into which range of the spreadsheet form, based on the input text $\mathcal{X}$ and auxiliary source files $\mathcal{S}$. 
Formally:
\begin{equation}
    \mathcal{O} = \{e_1, e_2, \dots, e_N\}
\end{equation}
where each \textit{filled entry} $e_i = (\ell_i, v_i, R_i)$ specifies the value $v_i$ to be written into range $R_i$ for label identifier $\ell_i$. Labels not supported by $\mathcal{X}$ or $\mathcal{S}$ are assigned $v_i = \text{null}$. Applying $\mathcal{O}$ to the spreadsheet form $\mathcal{F}$ yields the final filled spreadsheet.
\end{definition}

\begin{example}[Duplicated Labels]
As illustrated in Figure~\ref{fig:intro_scenarios}, the business card order spreadsheet form contains two repeated labels ``Name''. They are identified as $\ell_1 = (\text{Name}, 1)$ and $\ell_2 = (\text{Name}, 2)$. According to the provided source files, the resulting filled entries would be 
\[
\begin{aligned}
    e_1 &= (\ell_1, \text{``Robert J. Miller''}, \text{\{F4\}}), \\
e_2 &= (\ell_2, \text{``Robert J. Miller, Ph.D.''}, \text{\{B20\}}).
\end{aligned}
\]
\end{example}

As shown in the examples, the spreadsheet form filling task is inherently complex and distinguished from standard information extraction. First, spreadsheet forms often do not expose explicit schema-level label--value mappings, requiring models to infer implicit structure from layout signals (e.g., spatial alignment and layout grouping) to correctly associate labels with their corresponding writable regions. Second, correct completion is often context-dependent: auxiliary contents $\mathcal{C}$, such as instructions, section headers, or formatting constraints, may affect both what should be filled and how it should be written. Third, spreadsheet form filling requires exact range grounding, since predicted values must be written into the correct cells or ranges; even minor misplacement may overwrite template elements (e.g., labels or auxiliary contents) or the structural organization of the spreadsheet form.

\subsection{Evaluation Criteria}

Let $e_i = (\ell_i, v_i, R_i)$ be a predicted entry and $e_i^* = (\ell_i^*, v_i^*, R_i^*)$ be the corresponding ground-truth entry. We evaluate correctness at the entry level. A predicted entry is considered correct if both the semantic value and the target range match the ground truth:

\begin{equation}
\operatorname{Correct}(e_i,e_i^*)
=
\mathbb{I}\!\left[
\operatorname{MatchValue}(v_i,v_i^*)
\land
(R_i=R_i^*)
\right].
\end{equation}

Here, $\text{MatchValue}$ allows fuzzy matching between semantically equivalent textual values. This is important,  because in spreadsheet form filling, semantic correctness is often more critical than exact string equality. For example, the same value may appear in different surface forms, such as alternative date formats (e.g., ``2024-03-15'' vs. ``March 15, 2024''). In addition, if a value spans multiple cells or lines, the contents of the corresponding range are concatenated before comparison.

%\section{Running Example}
%\label{app:running}

%The business-card order form illustrates three difficulties discussed in Section~\ref{sec:intro}: evidence from heterogeneous files, irregular label--slot relationships, and consistency across multiple regions. The header and card-printing sections should refer to the same person, while the account number must be distributed across designated subfields. In the illustrated outputs, the VLM baseline leaves the approval field empty, a general agent selects the wrong Title slot, and a spreadsheet agent writes the account number into one cell. The worked examples in Appendix~\ref{app:implementation} trace the corresponding preprocessing and repair decisions.

\section{Pilot Study Details}
\label{app:pilot}

Before presenting our method, we conduct a pilot study to isolate the dominant failure patterns of general-purpose LLMs in spreadsheet form filling. To evaluate how representative systems handle structural grounding without conflating it with broader task complexity, we build a controlled pilot set. We randomly select 15 forms from two common structural categories (\emph{Mixed-Slot-Layout} and \emph{Repeated-Section Forms}) and manually expand them into 57 instances by keeping the filling semantics fixed while varying the relative spatial position between each label and its target slot.

\noindent$\bullet$ \bfit{Sensitivity to Relative Label--Slot Layouts.}
Figure~\ref{fig:pilot-study} shows that label--slot direction affects performance, with different sensitivity across systems. The left-label/right-slot arrangement is generally favorable, but OpenAI Agents SDK also performs strongly when the slot is below the label (0.842, compared with 0.858 on right-slot layouts). The pilot motivates attention to local correspondence rather than a claim that every noncanonical layout is difficult. 

\noindent$\bullet$ \bfit{Unstable but Structured Error Patterns.}
Beyond aggregate accuracy drops, we observe that failed outputs exhibit consistent, structured qualitative patterns. A single incorrectly filled form often contains multiple local errors concentrated in nearby but structurally incorrect regions. Common failure modes include writing directly into the label cell itself, filling an adjacent cell on the wrong side of the label, or generating locally plausible but role-inconsistent entries. Furthermore, repeated executions on the same form yield unstable outputs, shifting the error locations across runs. 

In summary, reliable spreadsheet form filling cannot be treated as unconstrained one-shot generation. The observed layout sensitivity and structured error patterns motivate a localized repair process that explicitly validates edits against template structure and re-grounds content using local geometric constraints.

\section{Implementation Details and Worked Examples}
\label{app:implementation}

\subsection{Form Filling Preprocessing}
\label{app:sec_preprocessing}

Existing systems often flatten structured inputs into one-dimensional sequences, which can obscure spatial relationships and weaken downstream reasoning over tables and form-like artifacts~\citep{xu2020layoutlm, Li0W0025}. Moreover, the evidence required for form filling is often scattered across heterogeneous auxiliary sources, such as emails, PDFs, spreadsheets, and other semi-structured records. 
Existing general-purpose agent frameworks~\citep{langgraph} and spreadsheet agents~\citep{li2023sheetcopilot} do not explicitly provide a unified representation that preserves the source-specific structural and spatial cues required in our form-filling setting.
% Existing general-purpose agent frameworks~\citep{langgraph} and spreadsheet agents~\citep{li2023sheetcopilot} are not well suited to this setting, as they largely flatten these inputs and fail to preserve the source-specific structural and semantic cues needed for reliable grounding. 
The core challenge is therefore to convert heterogeneous auxiliary sources into a unified yet structure-aware evidence representation that supports downstream form filling. To address this challenge, we design a preprocessing procedure that (1) represents the template spreadsheet form to provide an executable space and a layout-aligned visual modality enhancement, (2) transforms auxiliary sources into structured evidence representations and (3) performs an initial form pre-filling, thereby laying the foundation for subsequent refinement and precise re-filling.

\subsubsection{Dual-View Spreadsheet Representation}
\label{app:subsec_preprocessing_workspace}

\oursys first parses the spreadsheet template to recover cell coordinates, and determines the exact form range from the maximal populated row--column extent. Then, \oursys derives two complementary views over this range:

\noindent$\bullet$ \textbf{(1) Executable Spreadsheet Representation.} 
Over the identified form range, \oursys extracts both semantic and formatting attributes (e.g., cell content, fill color, and font properties), and constructs an executable spreadsheet representation from the raw spreadsheet data. In addition, \oursys derives a structural summary that captures merged-cell regions, writable and non-writable areas, and form-specific constraints such as dropdown validations together with their admissible candidate values. \textit{This representation enables the system to operate directly on constrained spreadsheet regions instead of an ambiguous flattened text view.}

\noindent$\bullet$ \textbf{(2) Layout Information Enhancement.} Because the executable spreadsheet representation mainly captures the form as a one-dimensional programmatic structure, it is insufficient for later reflection and validation stages that require explicit layout understanding, such as layout consistency checking and suspicious-range localization. To support these later stages, \oursys further augments the spreadsheet form with a layout view. Specifically, it first parses the spreadsheet into layout-aligned structured markup, then renders this markup into a sheet image, and records the alignment between spreadsheet coordinates, markup regions, and rendered regions. \textit{This layout-aligned representation preserves the layout organization of the form and links each cell region to its corresponding layout area.}

\subsubsection{Source File Preprocessing}
\label{app:subsec_preprocessing_sources}

To better preserve the semantic and structural information in source files for downstream form filling, \oursys preprocesses these source files in three steps:

\noindent$\bullet$ \textbf{(1) Type-Based Content Extraction.} \oursys first processes each file independently according to its type. (i) 
For \emph{text and markup} documents, the content is organized into chapters, with evidence spans preserved as textual grounding units. 
(ii) \emph{Structured files} (e.g., JSON, XML) are normalized into label--value records
, preserving the original label hierarchy whenever available.
(iii) \emph{Spreadsheets and tables} are represented by their column headers together with row-level records extracted from the table body.
(iv) \emph{Emails and HTML pages} are decomposed into header slots (sender, subject, date), body chapters, and embedded tables when present.
(v) \emph{PDFs and scanned files} are normalized into layout-organized text blocks, detected tables, and layout anchors used for later grounding.

\noindent$\bullet$ \textbf{(2) Structure-Based Evidence Formatting.} 
With the extracted file contents, \oursys further calculates a set of lightweight structural signals to characterize how the extracted content is organized, including table density, label--value regularity, chapter patterns, and repetition strength. Concretely, table density measures the fraction of extracted content tokens that belong to detected row--column regions among all extracted tokens in the file; label--value regularity measures how consistently text blocks can be paired into aligned label--value patterns with repeated relative offsets; section-pattern strength captures the frequency of title-like lines followed by grouped descendant content; and repetition strength captures the recurrence of records or rows sharing the same local schema.
\oursys then uses these signals to determine which structural pattern is most salient in the file and standardizes the extracted content accordingly. For example, (i) high label--value regularity yields a flat label--value view; (ii)  prominent chapter patterns induce sectioned blocks; (iii) strong row--column regularity or repeated-item structure leads to record- or table-oriented organization; (iv) deeply nested relations are preserved as hierarchical JSON-like structures; and (v) files exhibiting both discourse and tabular characteristics are represented in a hybrid chapter-and-table form. For weakly structured sources, \oursys retains evidence-grounded text spans as the primary representation.

\noindent$\bullet$ \textbf{(3) Source Bundle Assembling.} The processed results are then assembled into a unified source bundle that contains both normalized content and compact structural summaries, which is provided as a part of the prompt to the planner in the subsequent prefilling stage introduced in Section \ref{app:subsec_preprocessing_prefill}.

\subsubsection{Planner-Guided Prefilling}
\label{app:subsec_preprocessing_prefill}

With the carefully processed form template and source files, \oursys next utilizes a planner-guided prefill agent to conduct the initial form prefilling, whose output is further refined in the following stages (see Section~\ref{app:sec_srv}).

\noindent$\bullet$ \textbf{(1) Structure-Based Filling Plan Generation.} 
Given the structure-centered representation of the spreadsheet form extracted in Section~\ref{app:subsec_preprocessing_workspace}, \oursys first generates a structural summary by extracting key structural elements such as merged cells and label cells. Then we prompt the planning agent (by concatenating the summary, user instruction, source bundles) to  
generate a detailed prefilling plan: (1) It may specify planner-selected background reads that associate a target form region with the most relevant source bundles, or provide short operational notes to indicate how the region should be filled as shown in Figure~\ref{fig:framework}; (2) It also specifies global filling policies, such as whether unsupported or unspecified slots should be left blank according to the user’s instruction.

\noindent$\bullet$ \textbf{(2) Plan-Guided Prefill Execution.} The prefill agent receives the user instruction, the structured source bundle, the planner-derived policy, and the canonical form workspace as input. Under the generated prefilling plan, it fills the template spreadsheet form programmatically using the provided Excel skills.

To keep our prefill process scalable to relatively large spreadsheet forms, we adopt a fallback mechanism. When the canonical workbook is too large to be safely inlined into the model context, \oursys switches to a compact tool-first execution mode instead of forcing a full-context prompt. In this fallback mode, the agent is instructed to first inspect the workbook structure itself, and then read only small local ranges that are directly relevant to the current filling decision.

\subsection{Reflective Region Adjudication}
\label{app:sec_srv}

The prefilled form often contains many errors, such as overwriting labels instead of their corresponding slots, placing values into adjacent but incorrect cells, or violating form-specific constraints (e.g., dropdown validation rules). These errors are often difficult to detect because many incorrect filling actions appear plausible in isolation. For instance, a value written into a neighboring cell may seem reasonable, while still breaking the intended local label--value alignment. Such errors can substantially degrade downstream form-filling accuracy~\citep{stechly2024self,huang2023large}. To address this issue, in this section we introduce (1) a local reflective, evidence-grounded validation mechanism that identifies reliable and suspicious cells, along with (2) a range construction procedure that further consolidates them into compact suspicious regions for downstream repair.

\subsubsection{Evidence-Grounded Filling Validation}

To support reliable correction, we first perform an evidence-grounded validation process to identify suspicious cells that likely require repair and reliable cells that need to be protected. Rather than relying on unconstrained LLM self-reflection \citep{stechly2024self, huang2023large}, this process is explicitly grounded in the dual-view spreadsheet workspace constructed during preprocessing. The \emph{layout-aligned view} supports layout consistency checking against the original template, while the \emph{executable spreadsheet representation} exposes structural cues such as writable regions, static template cells, and slot-level constraints. The goal of this stage is therefore to determine which edits are potentially problematic and to attach sufficient diagnostic evidence for downstream correction.

The design of this validation module is guided by two principles. \textit{(1) Dual-view-grounded validation.} Suspicious edits are assessed jointly from layout evidence and structural execution semantics, rather than from free-form model introspection alone. \textit{(2) Decoupled discovery and adjudication.} Reflective Region Adjudication adopts a two-stage pipeline: the first stage performs broad screening to identify potentially problematic regions with high recall, and the second stage conducts localized diagnosis to determine the exact error type and validation state of each flagged location. This staged design allows downstream refill to operate over explicit repair targets instead of blindly regenerating the workbook.

\noindent$\bullet$ \textbf{Stage 1: Layout-Grounded Global Screening.}
The first stage performs a coarse scan over the entire prefilled workbook to identify \emph{reliable} and \emph{suspicious} cells. To achieve this, \oursys uses the \emph{layout-aligned view} (Section~\ref{app:subsec_preprocessing_workspace}) to compare the pristine template snapshot with the current prefilled workbook. By feeding these aligned layout views, together with the user instruction and edit history, into a VLM, layout-grounded global screening assesses whether the overall organization has been disrupted, for example due to misplaced or missing values, abnormal overwrites, or inconsistent local alignment. 

\noindent$\bullet$ \textbf{Stage 2: Structure-Aware Local Adjudication.}
The second stage complements the coarse candidates identified in Stage~1 by performing localized diagnosis over the \emph{executable spreadsheet representation}. Its goal is to determine whether each non-empty cell pattern exhibits any typical error mode observed in our pilot study (Section~\ref{sec:pilot-study}).

Structure-Aware Local Adjudication evaluates suspicious locations using two categories of spreadsheet-specific risk signals. \textit{(1) Template-integrity risks} are checked programmatically to detect whether a generated value has overwritten a static template cell, such as a label, title, or header. \textit{(2) Layout-grounding risks} are assessed by the VLM to capture structurally implausible fills, such as inverted label--value ordering or abnormal separation between a label and its associated value region.
Representative risk signals are summarized in Table~\ref{tab:srv-risk-signals}. A cell is marked as \emph{suspicious} only when it triggers at least two risk signals. This conservative threshold is intended to limit repairs triggered by a single noisy cue. As described in Section~\ref{sec:srv}, global screening can also contribute a repair candidate.

\begin{table}[t]
\centering
\small
\caption{Representative risk signals used in structure-aware local adjudication. These examples illustrate template-integrity and layout-grounding risks.}
\label{tab:srv-risk-signals}
\begin{tabular}{@{}p{0.25\linewidth}p{\dimexpr0.75\linewidth-2\tabcolsep\relax}@{}}
\hline
\textbf{Risk Signal} & \textbf{Illustrative Example} \\
\hline
\begin{tabular}[t]{@{}l@{}}\textbf{Label-cell}\\ \textbf{overwrite}\end{tabular} & The value ``11-2045-5-601002-4310'' is written into merged range \texttt{K19:T19}, which originally contains the static template label ``Account Number'', thereby overwriting template-side content.
 \\
\hline
\begin{tabular}[t]{@{}l@{}}\textbf{Semantic}\\ \textbf{ambiguity}\end{tabular} & The value ``Robert J. Miller'' is written into cell \texttt{B20}, even though this slot should be associated with the first occurrence of \texttt{Name} rather than the second one, creating ambiguity between duplicated label groups. \\
\hline
\begin{tabular}[t]{@{}l@{}}\textbf{Template}\\ \textbf{inconsistency}\end{tabular} & Cell \texttt{K20} is left-aligned, while the neighboring filled cells in the same local group are all center-aligned, indicating a local formatting inconsistency. \\
\hline
\begin{tabular}[t]{@{}l@{}l@{}}\textbf{Label--value}\\\textbf{layout}\\ \textbf{abnormality}\end{tabular} & The slot for \texttt{Signature} is located above its label, but the value is written into a nearby cell below or beside the label, violating local top-slot / bottom-label layout. \\

\hline
\end{tabular}
\end{table}

\subsubsection{Repair Range Construction}
\label{app:subsec_srv_ranges}

The identified \textit{suspicious} cells are not passed downstream in isolation. Instead, \oursys expands and consolidates them into compact \emph{suspicious ranges}, so that subsequent correction operates over coherent local regions rather than individual cells. This design avoids two undesirable extremes: repairing single cells would discard the surrounding layout context needed for re-grounding and refill, while regenerating the entire form would unnecessarily rewrite reliable regions and amplify error propagation.

Concretely, after combining the cell-level judgments from layout-grounded global screening and structure-aware local adjudication, \oursys expands suspicious cells in two stages to construct suspicious ranges. It first expands them in spreadsheet space, where merged cells are treated as atomic cells, so as to preserve sufficient local semantic and structural context. It then further expands the corresponding rendered regions in pixel space with a small margin, so that cell-border information and surrounding layout cues are also retained for downstream repair.

Each repair range further carries its adjudication evidence, including the detected error types and whether the range is associated with explicit template-integrity violations (e.g., \texttt{label-cell overwrite}), other structure-aware risk signals, or only a coarse warning from the global screening stage. This information directly determines the downstream repair strategy. For example, if Stage~2 reports a template-integrity risk, the refill module first triggers \emph{template restoration} before attempting to write new values. In contrast, ranges associated only with layout-grounding risks or content anomalies may require only localized refilling.

Finally, the planner determines whether each constructed suspicious range should enter the repair stage. Together, these outputs form a structured repair specification consisting of localized repair ranges, their diagnostic provenance, restoration indicators when needed, reliable cells, and cell-level validation states, which serves as the interface to the downstream repair module.

\subsection{Progressive Slot Re-grounding and Repair}
\label{app:sec_guided_refill}

Although Reflective Region Adjudication identifies suspicious ranges, reliable repair still requires a controlled local correction process rather than ordinary VLM-based refill within those ranges. The system must determine whether overwritten local template structure should first be restored, where revised values should be written, what nearby content must be preserved, and how local updates are applied without disturbing the surrounding structure. To address this, we introduce the \textit{Guided Refill} module, which turns prior validation outputs into constrained local repair actions while preserving reliable context. As shown in Figure \ref{fig:framework}, Guided Refill proceeds in four stages.

\noindent$\bullet$ \textbf{Step 1: Template Restoration.} Before guided refill, some suspicious ranges may need \emph{template restoration} if earlier execution has overwritten static template content. To detect this, \oursys compares the current mutable workbook against the immutable template snapshot at the cell level, checking cell values, cell styles, and merged-range structure. When overwriting is detected, \oursys restores the affected range to its original template state while preserving cells that have already been marked as reliable. This allows refill to proceed from a clean structural context and helps prevent earlier structural overwriting from propagating into later updates.

\noindent$\bullet$ \textbf{Step 2: Rectangle-Aware Slot Grounding.} After template restoration, local repair still requires accurate grounding of the writable slots within each suspicious range. This remains difficult for general-purpose vision-language agents, which have limited reliability in precise slot localization and require dedicated localization assistance~\citep{toles2026formgym}. We therefore introduce a dedicated pretrained model, the \emph{Rectangle-Aware Slot Grounder (RASG)}, to provide more stable candidate writable slots for subsequent local refill.

\hi{Model Design.} RASG is designed to localize writable cells by exploiting the rectangular geometry commonly exhibited by form slots. As illustrated in Figure~\ref{fig:RASG}, it adopts a lightweight SegFormer-B1~\citep{xie2021segformer} backbone with three task-specific prediction heads. Specifically, given the layout view of a spreadsheet form, RASG first encodes the image through four hierarchical stages. The resulting multi-scale features are then linearly projected, upsampled, and concatenated into a shared decoder representation for all three heads. The main head predicts the final rectangle logits, while the two auxiliary heads predict horizontal and vertical boundary cues, respectively. The rectangle logits are then rectified and reconstructed using the boundary predictions with a set of empirically chosen parameters. Finally, the reconstructed rectangle masks are aligned with candidate writable cells using area-based thresholds. By explicitly modeling cell-like rectangular geometry, RASG narrows the repair space to plausible writable slots and provides more task-aligned grounding for subsequent local refill.

\hi{Training Strategy.} RASG is trained with a three-stage strategy, which progressively increases data realism while preserving high slot recall. First, we train on synthetic spreadsheet forms with both bounding-box- and cell-level supervision, which allows RASG to learn basic label--slot layout patterns. In the second stage, we continue training on CommonForms~\citep{barrow2026commonforms}, a PDF-based dataset that provides only bounding-box supervision. This stage exposes RASG to more realistic rendering noise and layout variation, improving its robustness under weaker supervision. In the third stage, we further adapt the model on a manually filtered and annotated subset of FUSE~\citep{barik2015fuse}, which provides both bounding-box- and cell-level supervision for real-world spreadsheet forms. 

When no reliable slot is available, the controller falls back to guarded local repair using only the suspicious range, the template summary, and the protection policy.

\noindent$\bullet$ \textbf{Step 3: Local Label--Value Matching.} After RASG proposes a set of candidate writable cells, this stage converts these candidates into structured refill hints for subsequent local repair. In particular, we use a VLM to determine whether each candidate is a valid value cell and, if so, which nearby label it should be associated with.

Candidates judged valid are converted into positive hints that identify plausible value cells for a label, while invalid candidates are converted into negative hints that explicitly rule out non-target locations. In addition, \oursys may retain weaker context-only hints that preserve nearby label context when a specific target cell cannot yet be selected. Together, these hints turn local refill from unconstrained generation into a more guided local repair process, narrowing the feasible write space while preserving useful local context under ambiguity.

\noindent$\bullet$ \textbf{Step 4: Guided Refill Execution.} Using the original prompt, the range-level refill hints, the restored local spreadsheet state, and feedback from previous validation and repair steps, \oursys first constructs an explicit local refill plan that specifies where repair is allowed, which locations must be reconsidered, and which cells must remain protected. Under this plan and guided by the parsed hints, the refill agent performs local refill by generating and applying spreadsheet-editing actions to suspicious cells while leaving protected cells unchanged. 

\subsection{Qualitative Visualization of RASG}
\label{app:rasg_visualization}

\begin{figure*}[t]
    \centering
    \includegraphics[width=\textwidth]
    {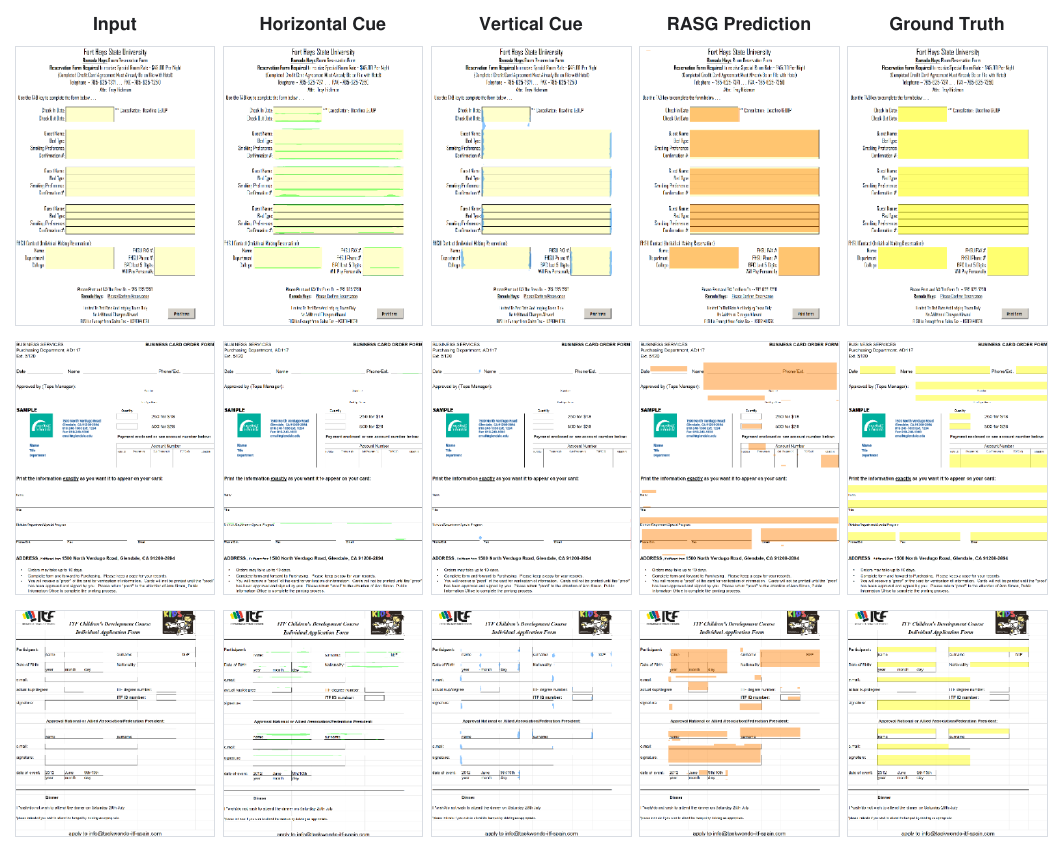}
    \caption{
    Qualitative visualization of RASG on representative spreadsheet
    forms. From left to right, we show the rendered input, horizontal
    boundary cues, vertical boundary cues, reconstructed RASG
    predictions, and ground-truth writable regions.
    }
    \label{fig:rasg_visualization}
\end{figure*}

Figure~\ref{fig:rasg_visualization} visualizes the intermediate
directional cues and final slot predictions produced by RASG on
representative form layouts. The horizontal and vertical cues capture
complementary slot boundaries, which are subsequently combined with
the segmentation output to reconstruct rectangular writable regions.
The examples illustrate how RASG represents candidate slots under
diverse layouts, including adjacent fields, elongated regions, and
irregularly arranged form sections.

\section{Benchmark Construction and Supplementary Metrics}
\label{app:benchmark}

Despite rapid progress in LLM-based agents for document interaction and spreadsheet automation~\citep{lin2025querying, chen2025sheetagent}, existing form filling benchmarks mainly target web-based or document-centered settings~\citep{li2025formfactory, toles2026formgym}, while existing spreadsheet evaluation efforts focus on spreadsheet understanding and manipulation rather than spreadsheet form filling~\citep{rahman2020benchmarking}. To our knowledge, spreadsheet form filling has not been explicitly studied as a dedicated benchmark task. To address this gap, we introduce \ourbench.

\hi{Benchmark Overview.} \ourbench is a benchmark for spreadsheet form filling, organized into two tracks to balance comparability and realism. \bfit{(1) Instruction-Only Track.} In this controlled setting, an agent fills a spreadsheet form from a natural-language instruction alone. \bfit{(2) Multi-File Track.} In this more realistic setting, an agent must consolidate evidence from one or more supporting files under a standardized prompt and complete the target form correctly. 

\begin{figure}[!t]
    \centering
    \includegraphics[width=\linewidth]{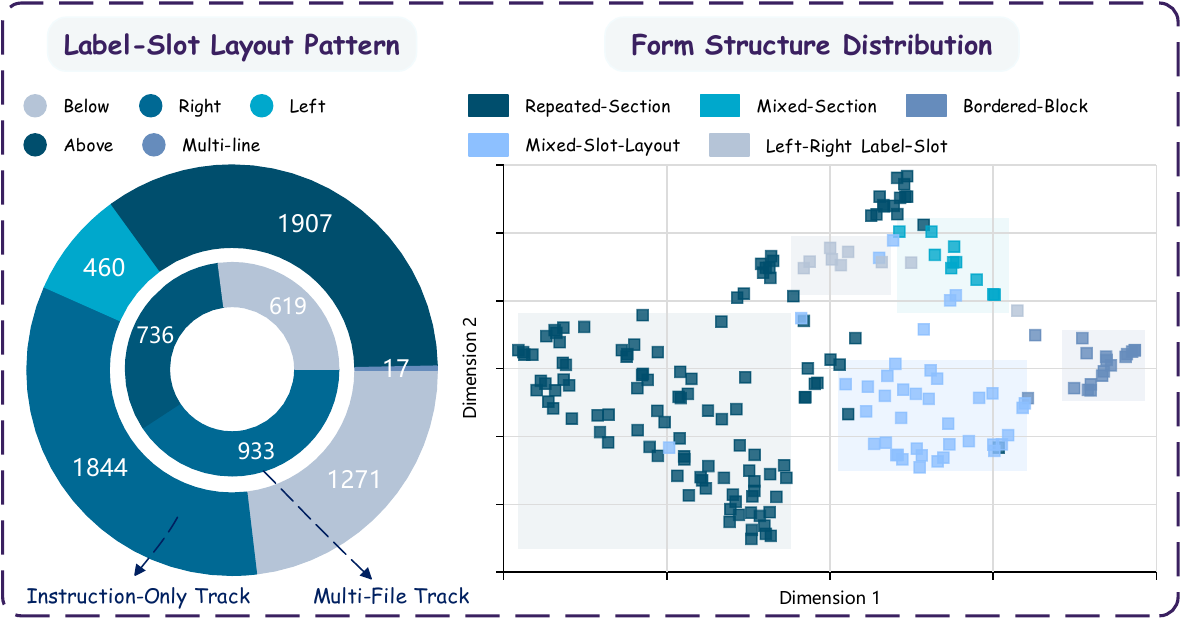}
    
    \caption{Statistics and Structural Diversity of \ourbench.}
    \label{fig:bench}
    
\end{figure}

\hi{Benchmark Construction.} The spreadsheet form templates are collected from FUSE~\citep{barik2015fuse}, VEUSES~\citep{xu2017spreadcluster}, and Venron2~\citep{xu2017spreadcluster, dou2016venron}, together with publicly collected online templates and additional synthetic templates generated with Claude Sonnet 4.6~\citep{anthropic2026sonnet46}. 

For the \bfit{Instruction-Only Track}, each spreadsheet form is paired with a synthesized natural-language instruction that specifies only the values to be filled, without revealing their target cell coordinates or explicit spatial relations in the form. Using the rendered form image and key annotations, we employ Gemini 3.1 Flash-Lite-Preview~\citep{google2026gemini31flashlite} to generate these value-only instructions.
All templates are normalized into a unified spreadsheet representation.
This track primarily emphasizes the spreadsheet-side form filling difficulty: even when the required information is already given in the instruction, agents may still fail to place values into the correct fields under complex and diverse spreadsheet layouts.

% For the  each spreadsheet form is paired with a synthesized natural-language instruction. 
% Using the rendered form image and key annotations, we employ Gemini 3.1 Flash-Lite-Preview to generate instructions describing the content to be filled. 

% For the Multi-File Track, all cases use a standardized instruction . The agent must instead identify and consolidate the required evidence from the accompanying source files and ground it to the target form.

For the \bfit{Multi-File Track}, each form is paired with several supporting files and a standardized prompt (e.g., “Fill the form based on the provided source files.”), which provides no field values or spatial hints, requiring the agent to use only explicitly supported information. The supporting files are drawn from open-source resources and synthetic enterprise document collections, spanning diverse departments, business scenarios, and heterogeneous file types such as emails, tables, notes, and semi-structured records. 
Compared with the Instruction-Only Track, this track places more emphasis on finding, consolidating, and organizing evidence across heterogeneous files before grounding the retrieved values back into the target form.

Synthetic templates are used to broaden structural coverage rather than to replace real-source forms. We therefore use them as a complement to public and corpus-derived templates, and all candidate samples undergo automated consistency checks followed by manual verification before inclusion.

\hi{Benchmark Statistics.} \ourbench contains 327 form instances in total: Among them, 200 samples with 5,499 label--value pairs belong to the \bfit{Instruction-Only Track}, while 127 samples with 2,288 pairs belong to the \bfit{Multi-File Track}. The \bfit{Multi-File Track} spans 11 file types, dominated by TXT (177), CSV (154), Markdown (103), JSON (95), and EML (80), with a longer tail, including YAML, presentation files, spreadsheets, images, and PDF files. Each Multi-File case is typically accompanied by around five supporting files.

Beyond source diversity, \ourbench also covers diverse local label--slot layouts, including \emph{above}, \emph{below}, \emph{left}, \emph{right}, and \emph{multi-line} patterns. As shown in Figure~\ref{fig:bench}, both tracks exhibit substantial layout diversity, although the \bfit{Multi-File Track} is more concentrated in \emph{above}, \emph{below}, and \emph{left} patterns.

\hi{Template-Level Structural Diversity.} To summarize template-level structural diversity in \ourbench, we derive five coarse archetypes from feature-based clustering over layout characteristics. \textit{Left-Right Label--Slot Forms} follow a canonical left-label/right-value arrangement for most fields. \textit{Mixed-Slot-Layout Forms} remain organized around label--value fields, but the relative placement of labels and values varies across regions. \textit{Repeated-Section Forms} consist of a vertical sequence of sections that repeat similar field-entry patterns. \textit{Mixed-Section Forms} combine several distinct sections with different internal layouts across the page. \textit{Bordered-Block Forms} are organized around prominent bordered or table-like blocks. These archetypes are intended as an interpretable summary of benchmark-level structural diversity rather than a strict taxonomy.

\hi{Evaluation Metrics.} We evaluate form-filling performance using two common metrics across both tracks, together with track-specific supplementary metrics. 

\noindent\bfit{(1) Pair Accuracy.} Pair Accuracy measures the proportion of target pairs for which both the filled value and the target cell or range are correct. It provides a fine-grained view of form-filling quality. 

\noindent\bfit{(2) End-to-End Accuracy.} End-to-End Accuracy measures the fraction of forms for which every target pair is correct, including fields that should remain blank. It is therefore a strict form-level metric. 

\noindent\bfit{(3) Fill Precision, Recall, and F1 for the Instruction-Only Track.} To distinguish underfilling from overfilling, we additionally evaluate the fill-versus-blank decision for each target field, independent of value correctness. We report Fill Precision, Recall, and F1 based on true positives (non-null fields that are filled), false negatives (non-null fields left blank), and false positives (null fields that are nevertheless filled).

\noindent\bfit{(4) Fill Accuracy for the Multi-File Track.} Since all target fields in the Multi-File Track are intended to be filled, we additionally report Fill Accuracy to measure the fraction of target fields that are filled by the system, regardless of value correctness.

\section{Supplementary Results and Analysis}
\label{app:analysis}
\subsection{Complete Benchmark Scores}
Table~\ref{tab:full_results} complements the primary metrics in Table~\ref{tab:main_results} with fill precision, recall, F1, and Fill Accuracy, following the definitions in Appendix~\ref{app:benchmark}.
\input{tables/full_results}
\begin{figure}[t]
    
    \centering  \includegraphics[width=0.95\linewidth]{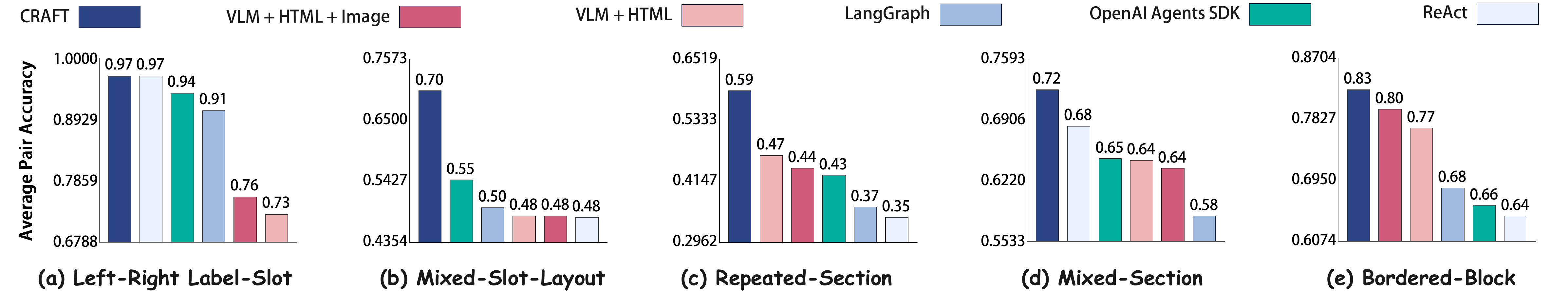}
    \caption{Comparison of average pair accuracy across five representative form types for \oursys and the top-5 baselines. The reported scores are computed by first measuring pair accuracy on individual forms and then averaging within each form type.}
    \label{fig:bar}
\end{figure}\subsection{Evaluation Across Spreadsheet Form Types} 
Figure~\ref{fig:bar} reports average pair accuracy across five representative form types. \oursys is top-performing across the five categories; on Left-Right Label--Slot Forms, it ties ReAct at the displayed precision (0.97). More importantly, these five types exhibit three distinct patterns.

\noindent\bfit{(1) Repeated-Section Forms and Mixed-Slot-Layout Forms.}
The advantage of \oursys is most pronounced on \emph{Repeated-Section Forms} and \emph{Mixed-Slot-Layout Forms}. On repeated-section forms, \oursys achieves 0.59, clearly above the strongest baseline at 0.47. On mixed-slot-layout forms, the gap is even larger, with \oursys reaching 0.70 while the best competing method remains at 0.55. These gains are consistent with the intended role of structure-aware validation and re-grounding: maintaining decisions across repeated units and recovering targets under noncanonical alignments. The aggregate subtype comparison does not isolate their individual effects.

\noindent\bfit{(2) Mixed-Section Forms.}
\oursys also remains the best-performing method on \emph{Mixed-Section Forms}. The main challenge here is that a single form contains multiple local sections with different internal organizations, so the system cannot rely on one dominant filling pattern and must adapt its decisions across sections. Although the gap is smaller than in the two categories above, the result is consistent with the value of coordinating section-specific decisions.  

\noindent\bfit{(3) Bordered-Block Forms and Left-Right Label--Slot Forms.}
On \emph{Bordered-Block Forms} and especially \emph{Left-Right Label--Slot Forms}, the gap to strong baselines becomes smaller. These categories expose clearer and more regular structural cues, so even competing methods can recover much of the local grounding correctly, leading to relatively high accuracy overall. \oursys remains competitive on these regular layouts, while its larger margins occur on mixed-slot and repeated-section forms.

Overall, these results suggest that the main advantage of \oursys lies in handling structurally irregular forms, while still maintaining strong performance on more regular layouts.

\begin{findingbox}
    \textbf{(Larger Margins on Mixed and Repeated Layouts):} In this benchmark, \oursys shows particularly clear gains on Mixed-Slot-Layout and Repeated-Section Forms. These subtype results describe where the system performs well; they do not separate layout effects from field count or source complexity.
\end{findingbox}

\begin{table}[t]
\centering
\caption{Representative case studies across both tracks. Outcomes are success (\cmark), partial success (\pmark), or failure (\xmark). Case metadata are listed below; L/R/U/D/ML denotes the original layout distribution.}
\label{tab:case_study_instruction}
\small
\setlength{\tabcolsep}{4pt}
\begin{tabular}{@{}lccccc|ccc@{}}
\toprule
& \multicolumn{5}{c|}{\textbf{Instruction-Only}} & \multicolumn{3}{c}{\textbf{Multi-File}} \\
\textbf{Method} & \textbf{4} & \textbf{29} & \textbf{38} & \textbf{84} & \textbf{88} & \textbf{226} & \textbf{238} & \textbf{284} \\
\midrule
VLM+HTML+Image & \cmark & \pmark & \cmark & \pmark & \cmark & \pmark & \pmark & \pmark \\
VLM+HTML & \pmark & \pmark & \cmark & \pmark & \cmark & \xmark & \pmark & \pmark \\
VLM+Excel Skills+Coding & \xmark & \xmark & \xmark & \xmark & \xmark & \xmark & \xmark & \xmark \\
\midrule
LangGraph & \cmark & \pmark & \cmark & \pmark & \cmark & \xmark & \pmark & \pmark \\
OpenAI Agents SDK & \cmark & \pmark & \cmark & \cmark & \pmark & \xmark & \pmark & \pmark \\
CrewAI & \pmark & \cmark & \cmark & \xmark & \cmark & \xmark & \pmark & \pmark \\
ReAct & \cmark & \pmark & \cmark & \cmark & \cmark & \xmark & \pmark & \pmark \\
\midrule
SheetAgent & \xmark & \xmark & \xmark & \xmark & \xmark & \xmark & \xmark & \xmark \\
SheetBrain & \xmark & \xmark & \xmark & \xmark & \xmark & \xmark & \xmark & \xmark \\
SheetCopilot & \xmark & \pmark & \pmark & \xmark & \cmark & \xmark & \xmark & \xmark \\
SheetMind & \xmark & \xmark & \xmark & \xmark & \xmark & \xmark & \xmark & \xmark \\
\midrule
\oursys & \cmark & \cmark & \cmark & \cmark & \cmark & \cmark & \cmark & \cmark \\
\bottomrule
\end{tabular}

\medskip
\begin{tabular}{@{}rlc@{}}
\toprule
\textbf{Task ID} & \textbf{Case Type} & \textbf{Layout (L/R/U/D/ML)} \\
\midrule
4 & Repeated-Section & 0/6/2/0/2 \\
29 & Mixed-Slot-Layout & 2/3/9/5/0 \\
38 & Left-Right Label--Slot & 0/6/2/0/0 \\
84 & Bordered-Block & 0/0/0/25/0 \\
88 & Mixed-Section & 0/11/0/0/0 \\
226 & Bordered-Block & 0/0/18/0/0 \\
238 & Mixed-Slot-Layout & 0/0/0/17/0 \\
284 & Repeated-Section & 0/20/0/0/0 \\
\bottomrule
\end{tabular}
\end{table}\subsection{Case Study}
\label{sec:case_study}

In Table~\ref{tab:case_study_instruction}, we compare representative cases from both the \bfit{Instruction-Only Track} and the \bfit{Multi-File Track}, and summarize each method's outcome as success, partial success, or failure. The selected cases highlight three recurring patterns. \textit{First}, vanilla VLM baselines remain competitive on visually regular single-form cases, such as Case 38, where at least one rendering-based baseline achieves success. However, they become much less reliable on structurally irregular layouts: for example, in Case 29, which involves a mixed-slot-layout, the vanilla VLM baselines achieve only partial success. \textit{Second}, spreadsheet-oriented agents show limited robustness beyond simpler spreadsheet-native settings. Although \textsc{SheetCopilot}~\citep{li2023sheetcopilot} succeeds in Case 88 and partially succeeds in Case 38, spreadsheet agents fail on most other structurally diverse cases, including Cases 4, 29, and 84. \textit{Third}, the \emph{Multi-File Track} remains challenging for nearly all baselines. On Cases 226, 238, and 284, most competing methods drop to partial success or failure, indicating difficulty in consolidating evidence across files and grounding the retrieved values back to the target form. Across all selected cases, \oursys is the only method that consistently achieves success.

%% file: tables/full_results.tex
\begin{table}[t]
\centering
\small
\caption{Complete benchmark results on the Instruction-Only and Multi-File tracks. All metrics are higher-is-better.}
\label{tab:full_results}
\setlength{\tabcolsep}{3.0pt}
\begin{adjustbox}{max width=\linewidth}
\begin{tabular}{lccccc|ccc}
\toprule
& \multicolumn{5}{c|}{\textbf{Instruction-Only Track}} & \multicolumn{3}{c}{\textbf{Multi-File Track}} \\
\cmidrule(lr){2-6} \cmidrule(lr){7-9}
\textbf{Method}
& \makecell{\textbf{Pair}\\\textbf{Acc. $\uparrow$}}
& \makecell{\textbf{E2E}\\\textbf{Acc. $\uparrow$}}
& \textbf{Rec. $\uparrow$}
& \textbf{Prec. $\uparrow$}
& \textbf{F1 $\uparrow$}
& \makecell{\textbf{Pair}\\\textbf{Acc. $\uparrow$}}
& \makecell{\textbf{E2E}\\\textbf{Acc. $\uparrow$}}
& \makecell{\textbf{Fill}\\\textbf{Acc. $\uparrow$}} \\

\midrule
\multicolumn{9}{l}{\textit{Vanilla VLM baselines}} \\
VLM+HTML+Image    &  \underline{51.17} & \underline{13.5} & 59.97 & 95.94 & \underline{73.80}& 32.60 & 12.60 &  39.25  \\
VLM+HTML          & 49.26 & 11.0 & 57.37 & \underline{95.96} & 71.80 
& 26.31 & 7.87 & 32.34  \\
VLM+Excel Skills+Coding  & 10.66 & 0.0 & \textbf{80.91} & 7.93 & 14.44 & 0.26 & 0.0 & 0.93  \\
\midrule
\multicolumn{9}{l}{\textit{General agent frameworks}} \\
LangGraph         & 44.28 & 10.0 & 56.19 & 95.54 & 70.76 & 33.65 & 12.60 & 40.82\\
OpenAI Agents SDK & 45.64 & 9.5 & 57.65 & 95.94 & 72.02 & \underline{43.14} & 16.54 & \underline{50.79} \\
CrewAI  & 41.77 & 9.5 & 51.96 & 95.73 & 67.36 & 40.25 & \underline{18.90} & 43.44 \\
ReAct  & 41.66 & 11.0 & 52.45 & 95.56 & 67.73 & 34.00 & 13.39 & 38.81 \\
\midrule
\multicolumn{9}{l}{\textit{Spreadsheet agents}} \\
SheetAgent  & 8.89 & 0.0  & 2.54  &58.41   & 4.87     & 0.26 & 0.0 & 0.00 \\
SheetBrain   & 10.98 & 0.0 & 3.58  &65.92   & 6.78     & 0.26 & 0.0 & 0.00 \\
SheetCopilot & 33.02 & 7.5   & 37.94  & 93.82   & 54.03     & 0.26 & 0.0 & 1.09 \\
SheetMind   & 10.69 & 0.0     & 3.13  &63.37   & 5.96     & 0.26 & 0.0 & 0.00 \\
\midrule
\multicolumn{9}{l}
{\textit{Ours}} \\
\oursys   & \textbf{59.68} & \textbf{15.5} & \underline{71.08} & \textbf{96.52} & \textbf{81.87}   & \textbf{66.52} & \textbf{43.31} & \textbf{90.87} \\
\bottomrule
\end{tabular}
\end{adjustbox}
\end{table}